\documentclass{article}

\usepackage{microtype}
\usepackage{graphicx}
\usepackage{subfigure}
\usepackage{booktabs} 
\usepackage{comment}
\usepackage[table,xcdraw]{xcolor}

\usepackage{hyperref}

\makeatletter
\@namedef{ver@algorithmic.sty}{}
\makeatother

\usepackage[accepted]{icml2020}

\usepackage{algorithm}
\usepackage[noend]{algpseudocode}
\usepackage{algorithmicx}

\usepackage{amsmath,amssymb,amsthm}
\usepackage{amsfonts}
\usepackage{amscd}
\usepackage{bbm}
\usepackage{mathrsfs}

\DeclareMathOperator*{\expected}{\mathbb{E}}

\newcommand{\statespace}{\mathcal{X}}
\newcommand{\actionspace}{\mathcal{A}}
\newcommand{\historyspace}{\ensuremath{\mathcal{H}}}
\newcommand{\I}{\ensuremath{\mathcal{I}}}

\newcommand{\rewardvar}{R}

\newcommand{\defi}{\ensuremath{\equiv}}
\newcommand{\indic}{\mathbbm{1}}

\newtheorem{definition}{Definition}

\newtheorem{property}{Property}

\icmltitlerunning{Temporally-Extended $\epsilon$-Greedy Exploration}

\excludecomment{longVersion}
\excludecomment{previousVersion}

\begin{document}

\twocolumn[
\icmltitle{Temporally-Extended $\epsilon$-Greedy Exploration}



\icmlsetsymbol{equal}{*}

\begin{icmlauthorlist}
\icmlauthor{Will Dabney}{dm}
\icmlauthor{Georg Ostrovski}{dm}
\icmlauthor{Andr\'e Barreto}{dm}
\end{icmlauthorlist}

\icmlaffiliation{dm}{DeepMind}

\icmlcorrespondingauthor{Will Dabney}{wdabney@google.com}

\icmlkeywords{Reinforcement learning, exploration, $\epsilon$-greedy}

\vskip 0.3in
]



\printAffiliationsAndNotice{}  

\begin{abstract}
Recent work on exploration in reinforcement learning (RL) has led to a series of increasingly complex solutions to the problem. This increase in complexity often comes at the expense of generality. Recent empirical studies suggest that, when applied to a broader set of domains, some sophisticated exploration methods are outperformed by simpler counterparts, such as $\epsilon$-greedy. In this paper we propose an exploration algorithm that retains the simplicity of $\epsilon$-greedy while reducing dithering. We build on a simple hypothesis: the main limitation of $\epsilon$-greedy exploration is its lack of temporal persistence, which limits its ability to escape local optima. We propose a temporally extended form of $\epsilon$-greedy that simply repeats the sampled action for a random duration. It turns out that, for many duration distributions, this suffices to improve exploration on a large set of domains. Interestingly, a class of distributions inspired by ecological models of animal foraging behaviour yields particularly strong performance.
\end{abstract}

\section{Introduction}\label{sec:intro}

Exploration is widely regarded as one of the most important open problems in reinforcement learning (RL). The problem has been theoretically analyzed under simplifying assumptions, providing reassurance and motivating the development of algorithms~\cite{brafman2002r,asmuth2009bayesian,azar2017minimax}. Recently, there has been considerable progress on the empirical side as well, with new methods that work in combination with powerful function approximators to perform well on challenging large-scale exploration problems~\cite{bellemare2016unifying,ostrovski2017count,burda2018exploration,badia2020never}. 

Despite all of the above, the most commonly used exploration strategies are still simple methods like  $\epsilon$-greedy, Boltzmann exploration and entropy regularization~\cite{peters2010relative,sutton2018reinforcement}. This is true for both work of a more investigative nature \cite{mnih2015human,silver2017mastering} and practical applications \cite{levine2016end,li2019transforming}. In particular, many recent successes of deep RL, from data-center cooling to Atari game playing, rely heavily upon these simple exploration strategies~\cite{mnih2015human,lazic2018data,kapturowski2018recurrent}.

Why does the RL community continue to rely on such naive exploration methods? There are several possible reasons. First, principled methods usually do not scale well. Second, the exploration problem is often formulated as a separate problem whose solution itself involves quite challenging steps. Moreover, besides having very limited theoretical grounding, practical methods are often complex and have significantly poorer performance outside a small set of domains they were specifically designed for. This last concern is perhaps the most severe, as an effective exploration method must be generally applicable. 
 
Naive exploration methods like $\epsilon$-greedy, Boltzmann exploration and entropy regularization are general because they do not rely on strong assumptions about the underlying domain. In part as a consequence of this, they are also \emph{simple}, not requiring too much implementation effort or per-domain fine tuning. This makes them appealing alternatives despite the fact that they may not be as efficient as some of their more complex counterparts. 

Maybe there is a middle ground between simple yet inefficient exploration strategies and more complex, though efficient, methods. The exploration method we propose in this paper represents a compromise between these two extremes. We ask the following question: how can we deviate as little as possible from the simple exploration strategies adopted in practice and still get clear benefits? In more pragmatic terms, we want a simple-to-implement algorithm that can be used in place of naive methods and lead to improved exploration.

In order to achieve our goal we propose a method that can be seen as a generalization of $\epsilon$-greedy---perhaps the simplest and most widely adopted exploration strategy. As is well known, in the $\epsilon$-greedy algorithm, at each time step the agent selects an exploratory action uniformly at random with probability $\epsilon$. Besides its simplicity, $\epsilon$-greedy has two properties that we believe contribute to its universality: 
 \begin{itemize}
    \item It is \emph{stationary}, i.e.\ its mechanics does not depend on learning progress. Stationarity is important for stability, since an exploration strategy interacting with the agent's learning dynamics results in circular dependencies that can in turn limit exploration progress. In simple terms: bad exploratory decisions can hurt the learned policy which can lead to more bad exploration.
    \item It provides full \emph{coverage} of the space of possible trajectories. All sequences of states, actions and rewards are possible under $\epsilon$-greedy exploration, albeit some with exceedingly small probability. This guarantees, at least in principle, that no solutions are excluded from consideration. Convergence results for classical RL algorithms rely on this sort of guarantee~\citep{singh2000convergence}. This may also explain why many sophisticated exploration methods still rely on $\epsilon$-greedy exploration~\citep{bellemare2016unifying,burda2018exploration}.
\end{itemize}

However, $\epsilon$-greedy in its original form also comes with drawbacks. Since it does not explore persistently, the likelihood of deviating more than a few steps off the default trajectory is vanishing small. This can be thought of as an inductive bias (or ``prior'') that favors transitions that are likely under the policy being learned (it might be instructive to think of a neighbourhood around the associated stationary distribution). Although this is not necessarily bad, it is not difficult to think of situations in which such an inductive bias may hinder learning. For example, it may be very difficult to move away from a local maximum if doing so requires large deviations from the current policy.

The issue above arises in part because $\epsilon$-greedy provides little flexibility to adjust the algorithm's inductive bias to the peculiarities of a given problem. By tuning the algorithm's only parameter, $\epsilon$, one can make deviations more or less likely, but the \emph{nature} of such deviations is not modifiable. To see this, note that all sequences of exploratory actions are equally likely under $\epsilon$-greedy, regardless of the specific value used for $\epsilon$. This leads to a coverage of the state space that is largely defined by the current (``greedy'') policy and the environment dynamics (see Figure~\ref{fig:intro} for an illustration).  

In this paper we present an algorithm that retains the beneficial properties of $\epsilon$-greedy while at the same time allowing for more control over the nature of the induced exploratory behavior. In order to achieve this, we propose a small modification to $\epsilon$-greedy: we replace actions with temporally-extended sequence of actions, or \emph{options}~\citep{sutton1999between}. Options then become a mechanism to modulate the inductive bias associated with $\epsilon$-greedy. We discuss how by appropriately defining a set of options one can ``align'' the exploratory behavior of $\epsilon$-greedy with a given environment or class of environments; we then show how a very simple set of domain-agnostic options work surprisingly well across a variety of well known environments.

\section{Background and Notation}
\label{sec:background}

Reinforcement learning can be set within the Markov Decision Process (MDP) formalism \cite{puterman1994markov}. An MDP $\mathcal{M}$ is defined by the tuple ($\statespace$, $\actionspace$, $P$, $\rewardvar$, $\gamma$), where $x \in \statespace$ is a state in the state space, $a \in \actionspace$ is an action in the action space, $P(x' \mid x, a)$ is the probability of transitioning from state $x$ to state $x'$ after taking action $a$, $R\colon \statespace \times \actionspace \to \mathbb{R}$ is the reward function and $\gamma \in [0, 1)$ is the discount factor. Let $\mathscr{P}(\actionspace)$ denote the space of probability distributions over actions; then, a policy $\pi\colon \statespace \to \mathscr{P}(\actionspace)$ assigns some probability to each action conditioned on a given state. We will denote by $\pi_a = \indic_a$ the policy which takes action $a$ deterministically in every state.

The agent attempts to learn a policy $\pi$ that maximizes the expected return or value in a given state,
{\small
\begin{equation*}
    V^\pi(x) = \mathbb{E}_{A \sim \pi} Q^\pi(x, A) = \mathbb{E}_{\pi} \left[ \sum_{t=0}^\infty \gamma^t R(X_t, A_t) \mid X_0 = x \right],
\end{equation*}}
\hspace{-2mm} where $V^\pi$ and $Q^\pi$ are the value and action-value functions of $\pi$. In this work we primarily rely upon methods based on the $Q$-learning algorithm \citep{watkins1992q}, which attempts to learn the optimal policy by approximating the Bellman optimality operator:
\begin{equation}
    \label{eq:qlearn}
	Q(x, a) = R(x, a) + \gamma \expected_{X' \sim P(\cdot \mid x, a)}\left[ \max_{a' \in \actionspace} Q(X', a') \right].
\end{equation}

In practice, the state space $\statespace$ is often too large to represent exactly and thus we have $Q_\theta(x, a) \approx Q(x, a)$ for a function approximator parameterized by $\theta$. We will generally use some form of differentiable function approximator $Q_\theta$, whether it be linear in a fixed set of basis functions (linear RL), or an artificial neural network (deep RL). We will then adjust the parameters $\theta$ by minimizing a squared (in linear RL) or Huber (in deep RL) loss function between the left- and right-hand sides of \eqref{eq:qlearn}, with the right-hand side held fixed \cite{riedmiller2005neural,mnih2015human}.

In addition to function approximation, it has been argued that in order to scale to large problems, RL agents should be able to reason at multiple temporal scales~\citep{dayan1993feudal,parr1998reinforcement,sutton1999between,dietterich2000hierarchical,hauskrecht2013hierarchical,kaelbling2014hierarchical}. One way to model temporal abstraction is through the concept of \emph{options}~\citep{sutton1999between}. Options are temporally-extended courses of actions. In their most general formulation, they can depend on the entire \emph{history} between time step $t$ when they were initiated and the current time step $t+k$, $h_{t:t+k} \defi x_t a_t x_{t+1}... a_{t+k-1} x_{t+k} $. Let \historyspace\ be the space of all possible histories; a \emph{semi-Markov option} is a tuple $\omega \defi (\I_\omega, \pi_\omega, \beta_\omega)$, where $\I_\omega \subset \statespace$ is the set of states where the option can be initiated, $\pi_\omega\colon \historyspace \to \mathscr{P}(\actionspace)$ is a policy over histories, and $\beta_\omega \colon \historyspace \mapsto [0,1]$ gives the probability that the option terminates after history $h$ has been observed~\citep{sutton1999between}. As in this work we will use options for exploration, we will assume that $\mathcal{I}_\omega = \statespace,\forall \omega$. 

Once an option $\omega$ is selected, the agent takes actions $a \sim \pi_\omega(\cdot \mid h)$ after having observed history $h$ and at each step terminates the option with probability $\beta_\omega(h)$. It is worth emphasizing that semi-Markov options depend on the history since their initiation, but not before. Also, they are usually defined with respect to a statistic of histories $h \in \historyspace$; for example, by looking at the \emph{length} of $h$ one can define an option that terminates after a fixed number of steps.

\section{Exploration in Reinforcement Learning}
\begin{figure}[t]
    \centering
    \includegraphics[width=.45\textwidth]{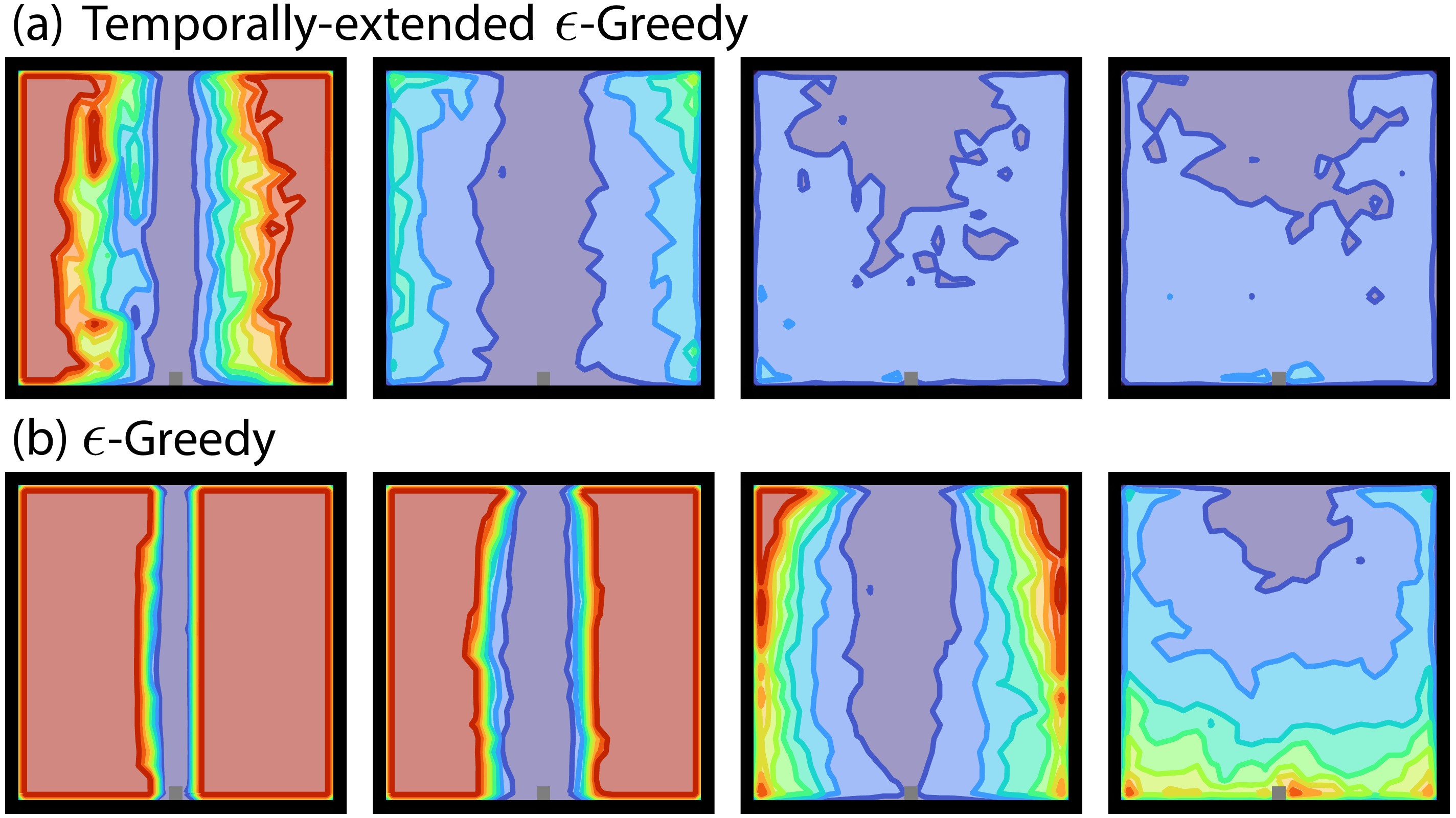}
    \vspace{-5px}
    \caption{Average, estimated from rollouts, first-visit times, comparing $\epsilon$-greedy policies with \textbf{(a)} and without \textbf{(b)} temporal persistence, in an open gridworld (blue represents fewer steps to first-visit and red states rarely or never seen). Greedy policy moves directly down from the start state (top center) until hitting the wall which ends the episode. See Appendix for full details.}
    \label{fig:intro}
    \vspace{-10px}
\end{figure}
At its core, RL presents the twin challenges of temporal credit assignment and exploration. The agent must accurately, and efficiently, assign credit to past actions for their role in achieving some long-term return. However, to continue improving the policy, the agent must also consider behaviours it (currently) estimates to be sub-optimal. This leads to the well-known \emph{exploration-exploitation trade-off}.

Because of its central importance in RL, exploration has been among the most extensively studied topics in the field. In finite state-action spaces, the theoretical limitations of exploration, with respect to sample complexity bounds, are fairly well understood \citep{azar2017minimax,dann2017unifying,agrawal2017optimistic}. However, these results are of limited practical use for two reasons. First, they bound sample complexity by the size of the state-action space, and horizon, which makes their immediate application in large-scale or continuous state problems difficult. Second, these algorithms tend to be designed based on worst-case scenarios, and can be inefficient on problems of actual interest.

Bayesian RL methods for exploration address the explore-exploit problem integrated with the estimation of the value-function itself \citep{kolter2009near,ghavamzadeh2015bayesian}. Generally such methods are strongly dependent upon the quality of their priors, which can be difficult to set appropriately. Thompson sampling based methods \cite{thompson1933likelihood,osband2013more} estimate the posterior distribution of value-functions, sample from this distribution and act greedily according to that sample. As with other methods which integrate learning and exploration into a single estimation problem, this creates non-stationary, but temporally persistent, exploration. Other examples of this type of exploration strategy include randomized prior functions \cite{osband2018randomized}, NoisyNets \cite{fortunato2017noisy}, parameter-space noise \cite{plappert2018parameter}, and successor uncertainties \cite{janz2019successor}. Although these methods are quite different from each other, they share key commonalities: non-stationary targets, temporal persistence, and exploration based on the space of value functions.

At the other end of the spectrum, there have recently been some successful attempts to design algorithms with specific problems of interest in mind.  This is true for example for a few games in the Atari-57 benchmark in which exploration is particularly challenging. Specifically, games such as \textsc{Montezuma's Revenge}, \textsc{Pitfall!}, \textsc{Private Eye}, and \textsc{Venture} have been identified as `hard exploration games' \cite{bellemare2016unifying}. This has attracted the attention of the research community and led to significant progress on these games in terms of performance \citep{ecoffet2019go,burda2018exploration}. On the downside, these results have been usually achieved by algorithms with little or no theoretical grounding, adopting specialized inductive biases, such as density modeling of images \cite{bellemare2016unifying,ostrovski2017count}, error-seeking intrinsic rewards \cite{pathak2017curiosity,burda2018exploration}, or perfect deterministic forward-models \cite{ecoffet2019go}. 

Generally, such algorithms are evaluated only on the very domains they are designed to perform well on, raising reasonable questions of generality. Recent empirical analysis sheds light on this matter by showing that some of these methods perform similarly to each other on challenging exploration problems and significantly under-perform $\epsilon$-greedy otherwise~\citep{taiga2020on}. One aspect of this is that complex algorithms tend to be more brittle and harder to reproduce, and thus lead to lower than expected performance in follow-on work. However, these results also strongly suggest that much of the recent work on exploration is over-fitting to a small number of domains.

\section{Temporally-Extended Exploration}

There are many ways to think about exploration: curiosity, experimentation, reducing uncertainty, etc. Consider viewing exploration as a search for undiscovered rewards or shorter paths to known rewards. In this context, the behaviour of uniform $\epsilon$-greedy appears terribly shortsighted because the probability of moving consistently in any way decays exponentially with the number of steps of exploration. In Figure~\ref{fig:intro}b we visualize the behaviour of uniform $\epsilon$-greedy in an open gridworld, where the agent starts at the center-top and the greedy policy moves straight down. Observe that for values of $\epsilon \le 0.5$ the agent is exceedingly unlikely to reach states outside a narrow band around the greedy policy. Even as the policy becomes purely exploratory ($\epsilon = 1.0$), the agent requires a large number of steps to ever visit the bottom corners of the grid. This is because, under the uniform policy, the probability of moving consistently in any direction decays exponentially (see Figure~\ref{fig:intro}b). By contrast, a method that explores persistently with a directed policy leads to more efficient exploration of the space at various values of $\epsilon$ (Figure~\ref{fig:intro}a).

The importance of temporally-extended exploration has been previously highlighted \citep{osband2016deep}, and in general, count-based \citep{bellemare2016unifying,ostrovski2017count,tang2017exploration} or curiosity-based \citep{pathak2017curiosity,burda2018exploration} exploration methods are inherently temporally-extended due to integrating exploration and exploitation into the greedy policy. Here our goal is to leverage the benefits of temporally-extended exploration without modifying the greedy policy.

In order to discuss temporally-extended exploration we will make use of the formalism of options introduced in Section~\ref{sec:background}. Options are a strict generalization of actions, so we can use the former to implement any exploration strategy based on the latter. For example, the uniform exploration used in Figure~\ref{fig:intro}b can be achieved by defining a uniform policy over the set of options $\omega_a \equiv (\statespace, {\pi}_a, \beta)$, where $\pi_a(h) = \indic_a$ and $\beta(h) = 1$ for all $h \in \historyspace$. 

There has been a wealth of research on learning options \cite{mcgovern2001automatic,stolle2002learning,csimcsek2004using,bacon2017option,jinnai2019finding,harutyunyan2019termination}, specifically for exploration \cite{2018eigenoption,jinnai2019discovering,jinnai2020exploration,Hansen2020Fast}. Often, these methods use options for both exploration and to directly augment the action-space, adding learned options to the actions available at states where they can be initiated. 

In the remainder of this work, we repeatedly argue for temporally-extended exploration, using options to encode a set of inductive biases to improve sample-efficiency. This fundamental message is found throughout the existing work on exploration with options, but producing algorithms that are empirically effective on large environments remains a challenge for the field. In the next section, we discuss in more detail how the options' policy $\pi_{\omega}$ and termination $\beta_{\omega}$ can be used to induce different types of exploration.

\subsection{Temporally-Extended $\epsilon$-Greedy}

A \emph{temporally-extended $\epsilon$-greedy} exploration strategy depends on choosing an exploration probability $\epsilon$, a set of options $\Omega$, and a sampling distribution $p$ with support $\Omega$. Then, on each step the agent follows the current policy $\pi$ for one step with probability $1-\epsilon$, or with probability $\epsilon$ samples an option $w \sim p(\Omega)$ and follows it until termination.

As discussed in the introduction, standard $\epsilon$-greedy has three desirable properties that help explain its wide adoption in practice: it is \emph{simple}, \emph{stationary}, and promotes full \emph{coverage} of the state-action space in the limit (which allows for convergence to the optimal policy under the right conditions). We now discuss to what extent the proposed algorithm retains these properties. Although somewhat subjective, it seems fair to call temporally-extended $\epsilon$-greedy a simple method. It is also stationary when the set of options $\Omega$, and distribution $p$, are fixed, for in this case its mechanics are not influenced by the data it collects. Finally, it is easy to define conditions under which temporally-extended $\epsilon$-greedy covers the entire state-action space, as we discuss next.

Obviously, the exploratory behavior of temporally-extended $\epsilon$-greedy will depend on the set of options $\Omega$. Ideally we want all actions $a \in \actionspace$ to have a nonzero probability of being executed in all states $x \in \statespace$ regardless of the greedy policy $\pi$. This is clearly not the case for all sets $\Omega$. In fact, this may not be the case even if for all $(x,a) \in \statespace \times \actionspace$ there is an option $\omega \in \Omega$ such that $\pi_{\omega}(a | hx) > 0$, where $hx$ represents all histories ending in $x$. To see why, note that, given a fixed $\Omega$ and $\epsilon > 0$, it may be impossible for an option $\omega \in \Omega$ to be ``active'' in state $x$ (that is, either start at or visit $x$). For example, if all options in $\Omega$ terminate after a fixed number of steps that is a multiple of $k$, temporally-extended $\epsilon$-greedy with $\epsilon = 1$ will only visit states of an unidirectional chain whose indices are also multiples of $k$. Perhaps even more subtle is the fact that, even if all options can be active at state $x$, the \emph{histories} $hx \in \historyspace$ associated with a given action $a$ may themselves not be realizable under the combination of $\Omega$ and the current $\pi$.

It is clear then that the coverage ability of temporally-extended $\epsilon$-greedy depends on the interaction between $\pi$, $\Omega$, $\epsilon$, and the dynamics $P(\cdot|x,a)$ of the MDP. One way to reason about this is to consider that, once fixed, these elements induce a stochastic process which in turn gives rise to a well-defined distribution over the space of histories \historyspace.

\begin{property}[Full coverage]
\label{prop:coverage}
Let $\mathcal{M}$ be the space of all MDPs with common state-action spaces $\statespace$,$\actionspace$, and $\Omega$ a set of options defined over this state-action space. Then, $\Omega$ has \textbf{full coverage} for $\mathcal{M}$ if $\forall M \in \mathcal{M}$, $\epsilon > 0$, and $\pi$, the semi-Markov policy $\mu := (1 - \epsilon) \pi + \epsilon \pi_\omega$, where $\omega$ is a random variable uniform over $\Omega$, visits every state-action pair with non-zero probability. Note that $\mu$ is itself a random variable and not an average policy.
\end{property}

We can then look for simple conditions that would lead to having Property~\ref{prop:coverage}. For example, if the options' policies only depend on the last state of the history, $\pi_{\omega}(\cdot|hx) = \pi_{\omega}(\cdot|x)$ (that is, they are Markov, rather than semi-Markov policies), we can get the desired coverage by having $\pi_{\omega}(a|x) > 0$ for all $x \in \statespace$ and all $a \in \actionspace$. The coverage of $\statespace \times \actionspace$ also trivially follows from having all primitive actions $a \in \actionspace$ as part of $\Omega$. Note that if the primitive actions are the \emph{only} elements of $\Omega$ we recover standard $\epsilon$-greedy, and thus coverage of $\statespace \times \actionspace$. Of course, in these and similar cases, temporally-extended $\epsilon$-greedy allows for the convergence to the optimal policy under the same conditions as its precursor.

Unfortunately, convergence in-theory is not enough for an efficient exploration algorithm in-practice. For sample-efficiency we want to cover the state-action space \textit{quickly}. However, unlike Property~\ref{prop:coverage}, how quickly coverage occurs will depend heavily on the alignment of the inductive biases of $p(\Omega)$ and a particular MDP. 

\begin{definition}[Cover time]
\label{def:covertime}
Let $M$ be an MDP, $\Omega$ a set of options with Property~\ref{prop:coverage} for $M$ and some $\epsilon > 0$ and $\pi$. The \textbf{cover time} of temporally-extended $\epsilon$-greedy with sampling distribution $p(\Omega)$ is the number of steps needed to visit all state-action pairs at least once with probability $1/2$ starting from the initial state distribution.
\end{definition}

\citet{even2003learning} show that the sample efficiency of Q-learning can be bounded in terms of the cover time of the exploratory policy. \citet{liu2018simple} extend this analysis to study the properties of MDPs for which uniform random exploration can be sample efficient. Meanwhile, \citet{jinnai2019discovering} provide an algorithm to learn point-options (transitioning from one state to one other state) that minimize cover time. Although challenging to analyze in full generality, to achieve efficient exploration on problems of interest we want to design our set of options and sampling distribution, $p(\Omega)$, to minimize the expected cover time on MDPs of interest while maintaining Property~\ref{prop:coverage} broadly.

This view of temporally-extended $\epsilon$-greedy as inducing a stochastic process also helps to understand its differences with respect to its standard counterpart. Since the induced stochastic process defines a distribution over histories we can also talk about distributions over sequences of actions. With standard $\epsilon$-greedy, every sequence of $k$ exploratory actions has a probability of occurrence of exactly $(\epsilon / |\actionspace|)^k$, where $|\actionspace|$ is the size of the action space. By changing $\epsilon$ one can uniformly change the probabilities of \emph{all} length-$k$ sequences of actions, but no sequence can be favored over the others. Temporally-extended $\epsilon$-greedy provides this flexibility through the definition of $\Omega$; specifically, by defining the appropriate set of options one can control the temporal correlation between actions. Changing this distribution over action sequences (histories), through our definition of $\Omega$, encodes the inductive bias of our exploration policy, directly affecting what types of MDPs will have low cover times. Next, we propose a concrete form of temporally-extended $\epsilon$-greedy which requires no specific domain knowledge but encodes a commonly held inductive bias: actions have (largely) consistent effects throughout the state-space.

\begin{figure*}[t]
    \centering
    \includegraphics[width=.75\textwidth]{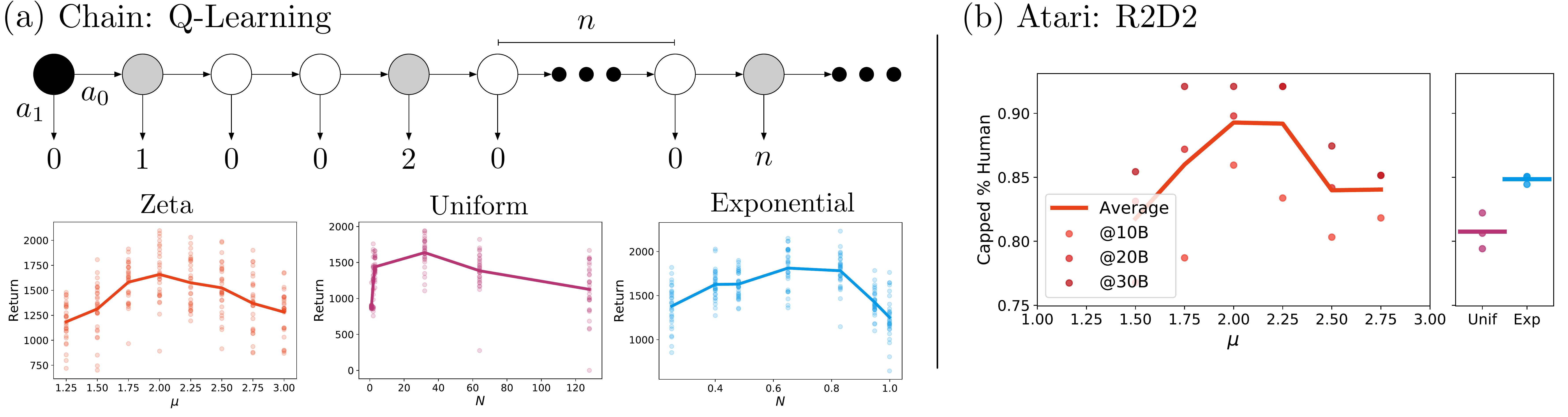}
    \vspace{-5px}
    \caption{\textbf{(a)} Modified chain MDP, action $a_0$ moves right, $a_1$ terminates with specified reward. Rewards follow a pattern of n zeros followed by a single reward $n$, etc. Evaluation of performance under various duration distributions and hyper-parameters on the chain. \textbf{(b)} Duration distribution similarly compared for an R2D2-based deep RL agent in Atari. }
    \label{fig:example_climb}
\end{figure*}

\subsection{$\epsilon z$-Greedy}
We begin with the options $\omega_a$ defined previously and consider a single modification, temporal persistence. Let $\omega_{an} \equiv (\statespace, \pi_a, \beta(h) = \indic_{|h| == n})$ be the option which takes action $a$ for $n$ steps and then terminates. Our proposed algorithm, as a first step towards temporally-extended $\epsilon$-greedy, is to let $\Omega  = \{\omega_{an} \}_{a\in \actionspace,n \ge 1}$ and $p$ to be uniform over actions with durations distributed according to some distribution $z$. Intuitively, we are proposing the set of semi-Markov options made up of all ``action-repeat'' policies for all combinations of actions and repeat durations, with a parametric sampling distribution on durations.

This exploration algorithm is then described by two parameters, $\epsilon$ dictating when/how often to explore, and $z$ dictating the degree of persistence. Notice that when $z$ puts all mass on $n = 1$, this is standard $\epsilon$-greedy, and more generally this combination of distributions forms a composite distribution with support $[0, \infty)$, which is to say that with some probability the agent explores for $n = 0$ steps, corresponding to following its usual policy, and for all other $n > 0$ the agent explores following an action-repeat policy.

A natural question arises: what distribution over durations should we use? To help motivate this question, and to help understand the desirable characteristics, consider Figure~\ref{fig:example_climb} which shows a modified chain MDP with two actions. Taking the `down' action immediately terminates with the specified reward, whereas taking the `right' action progresses to the next state in the chain. Similar to other exploration chain-like MDPs \cite{osband2018randomized}, $\epsilon$-greedy performs poorly here because the agent must move consistently in one direction for an arbitrary number of steps (determined by the discount factor) to reach the optimal reward. 

Instead, we consider the effects of three classes of distributions over duration: exponential ($z(n) \propto \lambda^{n-1}$), uniform ($z(n) = \indic_{n \le N} / N$), and zeta ($z(n) \propto n^{-\mu}$). Figure~\ref{fig:example_climb}b shows the average return achieved by these duration distributions as their hyper-parameters are varied. This problem illustrates that, without prior knowledge of the MDP, it is important to support arbitrarily long durations, such as with a heavy-tailed distribution like zeta. 

Why not simply allow uniform over an extremely large support? Doing so would effectively force `pure' exploration without any exploitation, because this form of \emph{ballistic} exploration would simply continue exploring indefinitely. Indeed, we can see this in both the results of the uniform distribution, and for zeta as $\mu \to 1$. On the other hand, short durations lead to frequent switching and vanishingly small probabilities of reaching larger rewards at all.

This type of trade-off leads to the existence of an optimal value of $\mu$ for the zeta distribution that can vary by domain \cite{humphries2010environmental}, and is illustrated by the inverted U-curve in Figure~\ref{fig:example_climb}. Interestingly, this finding is not unique to RL. A class of ecological models for animal foraging known as \emph{L\'evy flights} follow a similar pattern of choosing a direction uniformly at random, and following that direction for a duration sampled from a heavy-tailed distribution \cite{viswanathan1996levy,sims2012levy,baronchelli2013levy}. Under certain conditions, this has been shown to be an optimal foraging strategy, a form of exploration for a food source of unpredictable location \cite{viswanathan1999optimizing}. 

Thus, in the remainder of this work we will use the heavy-tailed zeta distribution, with $\mu = 2$ unless otherwise specified, and call this combination of $\epsilon$ chance to explore and zeta-distributed durations, $\epsilon z$-Greedy exploration\footnote{Pronounce `easy-greedy'.}.

\begin{figure*}[t]
    \centering
    \includegraphics[width=\textwidth]{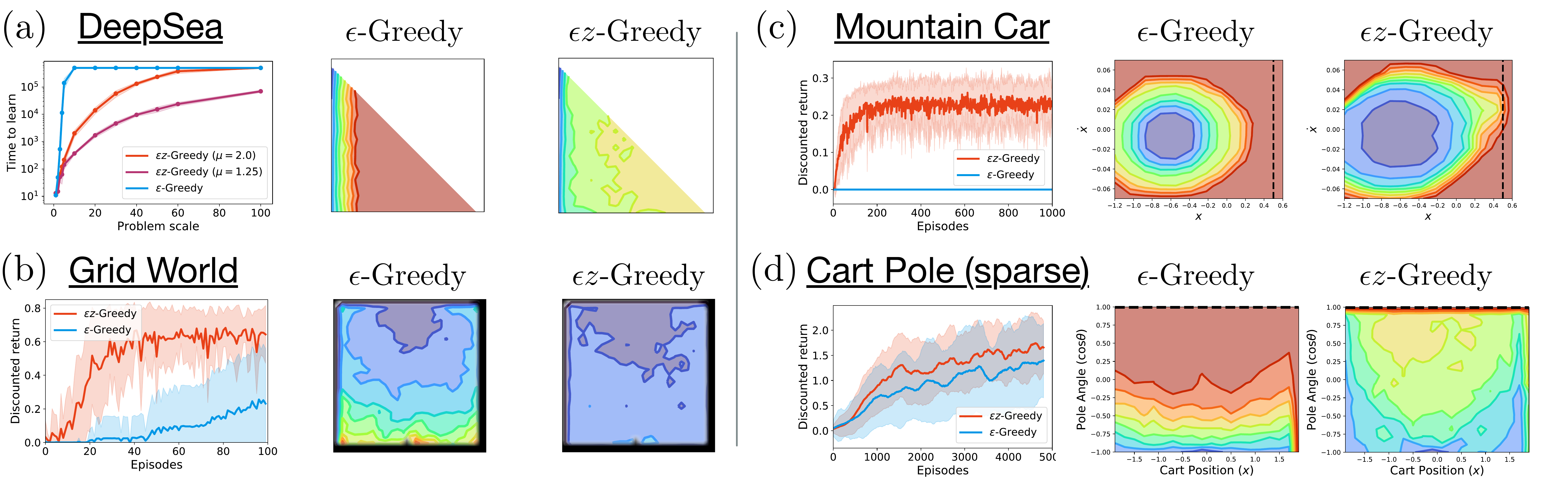}
    \vspace{-15px}
    \caption{Comparing $\epsilon$-greedy with $\epsilon z$-greedy on four small-scale domains requiring exploration. (a) DeepSea is a tabular problem in which only one action-sequence receives positive reward, and uniform exploration is exponentially inefficient, (b) GridWorld is a four-action gridworld with a single reward, (c) MountainCar is the sparse reward (only at goal) version of the classic RL domain, and (d) CartPole swingup-sparse only gives non-zero reward when the pole is perfectly balanced and the cart near-center. For each environment we show performance comparing $\epsilon$-greedy with $\epsilon z$-greedy (left), as well as average first-visit times over states for both algorithms during \textbf{pure exploration} ($\epsilon = 1$). In all first-visit plots, color levels are linearly scaled, except for DeepSea in which we use a logarithmic scale.}
    \label{fig:toy}
\end{figure*}

\section{Experimental Results}
We have emphasized the importance of simplicity, generality (via convergence guarantees), and stationarity of exploration strategies. We proposed a simple temporally-extended $\epsilon$-greedy algorithm, $\epsilon z$-greedy, and saw that a heavy-tailed duration distribution yielded the best trade-off between temporal persistence and sample efficiency. 

In this section, we present empirical results on tabular, linear, and deep RL settings, pursuing two objectives: The first is to demonstrate the generality of our method in applying it across domains as well as across multiple value-based reinforcement learning algorithms (Q-learning, SARSA, Rainbow, R2D2). Second, we make the point that exploration comes at a cost, and that $\epsilon z$-greedy improves exploration with significantly less loss in efficiency on dense-reward domains compared with existing exploration algorithms.

\subsection{Small-Scale Environments: Tabular \& Linear RL}

We consider four small-scale environments: DeepSea, GridWorld, MountainCar, and CartPole swingup-sparse. All four environments are configured to be challenging exploration problems with sparse rewards. We largely defer to referenced works for full details on each domain, but will highlight important aspects for each.

DeepSea is a needle-in-a-haystack 2-action chain problem where only a single action-sequence produces positive reward \cite{osband2018randomized}; we evaluate on the environment variant with consistent action effects in the main text and include the randomized action effects version in the appendix. GridWorld is a 4-action, $23\times23$ gridworld with fixed start state and a single non-zero reward on the other side of the room at a diagonal (see Appendix for details). MountainCar requires control of an under-powered car stuck in a valley to reach the top of the hill on one side, with three actions, and zero reward except for $1.0$ at the goal \cite{sutton2018reinforcement}. Finally, CartPole swing-up with sparse rewards is another physics-based control problem, with three actions controlling acceleration of a cart, where the agent must use momentum to swing the pole upright, and then balance it there, to receive any reward \cite{tassa2018deepmind}. For $\theta$ the pole angle and $x$ the cart position, the reward is zero unless $\cos(\theta) > 0.995$ and $|x| < 0.25$. DeepSea and GridWorld use a tabular representation while MountainCar and CartPole use linear function approximation on top of an order $5$ and $7$ Fourier basis respectively \cite{konidaris2011value}.

In Figure~\ref{fig:toy} we present results comparing $\epsilon$-greedy and $\epsilon z$-greedy on these four domains. Unless otherwise specified, the hyper-parameters and training settings for these two methods are identical. For each domain we show (i) learning curves showing average return against training episodes, (ii) average first-visit times on states during pure ($\epsilon = 1.0$) exploration for $\epsilon$-greedy and (iii) $\epsilon z$-greedy. 

The results show that $\epsilon z$-greedy provides significantly improved performance on these domains. The first-visit times provide some insight into this, showing significantly better coverage over the state-space compared with $\epsilon$-greedy.

\subsection{Atari-57: Deep RL}

\begin{figure*}[t]
    \centering
    \includegraphics[width=0.9\textwidth]{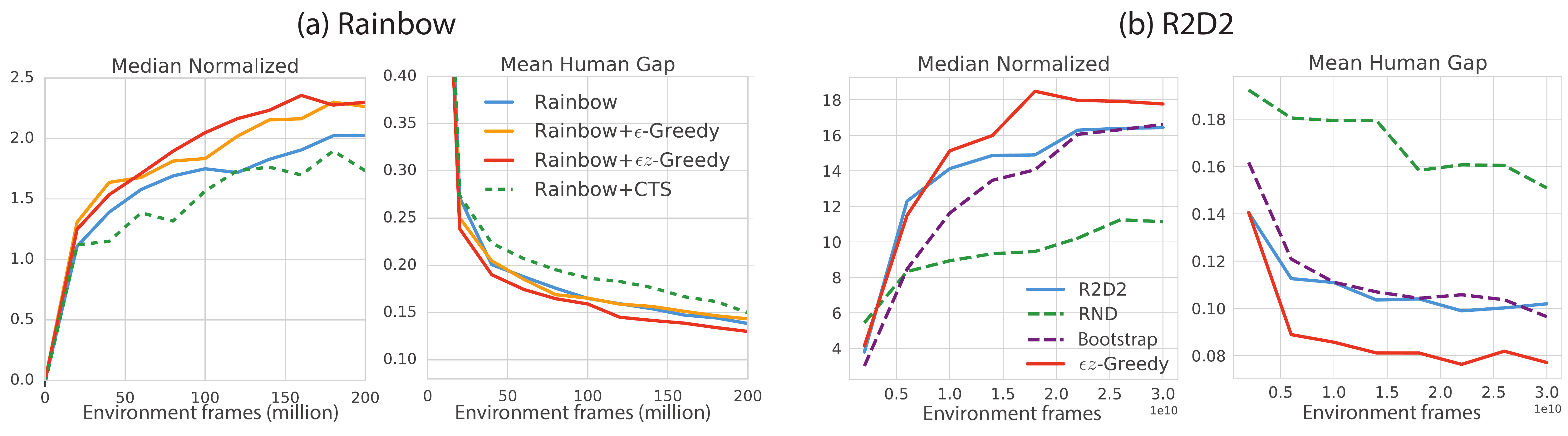}
    \vspace{-15px}
    \caption{Results on the Atari-57 benchmark for (a) Rainbow-based agents and (b) R2D2-based agents.}
    \label{fig:atari}
\end{figure*}
\begin{figure*}[t]
    \centering
    \includegraphics[width=\textwidth]{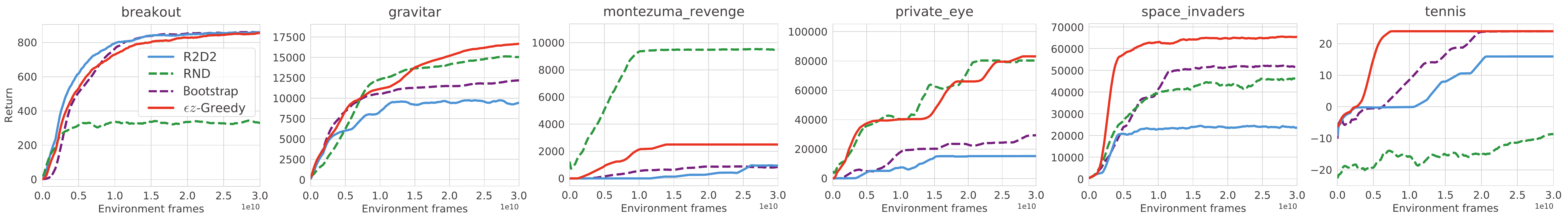}
    \vspace{-10px}
    \caption{Results on the Atari-57 selected games showing R2D2, R2D2 with $\epsilon z$-greedy, and R2D2 with RND exploration.}
    \label{fig:games}
\end{figure*}

Motivated by the results in tabular and linear settings, we now turn to deep RL and evaluate performance on 57 Atari 2600 games in the Arcade Learning Environment (ALE) \cite{bellemare2013arcade}. To demonstrate the generality of the approach, we apply $\epsilon z$-greedy to two state-of-the-art deep RL agents, Rainbow \cite{hessel2018rainbow} and R2D2 \cite{kapturowski2018recurrent}. We compare with baseline performance as well as the performance of a recent intrinsic motivation-based exploration algorithm: CTS-based pseudo-counts \cite{bellemare2016unifying} in Rainbow and RND \cite{burda2018exploration} in R2D2, each tuned for exploration performance comparable with published results. Finally, in R2D2 experiments we also compare with a Bootstrapped DQN version of R2D2 \citep{osband2016deep}, providing an exploration baseline without intrinsic rewards. We include full pseudo-code and hyper-parameters in the Appendix for reference, though the implementation of $\epsilon z$-greedy in each case is trivial, hyper-parameters are mostly identical to previous work, and we fix $\mu = 2$ for all results in this section.

Our findings (see Figure~\ref{fig:atari}) show that $\epsilon z$-greedy improves performance on the hard exploration tasks with little to no loss in performance on the rest of the suite. By comparison, we observe that the intrinsic motivation methods often (although not always) outperform $\epsilon z$-greedy on the hard exploration tasks, but at a significant loss of performance on the rest of the benchmark. 

The results in Figure~\ref{fig:atari} show median human-normalized score over the 57 games and the human-gap, which measures the degree to which the agent under-performs humans on average (see Appendix~\ref{app:expdetails} for details). We consider the median to indicate overall performance on the suite, although full per-game and mean performance is given in the Appendix, and the human-gap to illustrate gains on the hard exploration games where agents still under-perform relative to humans. In Figure~\ref{fig:games} we give representative examples of per-game performance for the R2D2-based agents. These per-game results make a strong point, that even on the hard exploration games the inductive biases of intrinsic motivation methods may be poorly aligned (see for example \textsc{Private Eye}), and that outside a small number of games these methods significantly hurt performance, whereas our proposed method improves exploration while avoiding this significant loss elsewhere.

Finally, we recall the ablation over duration distributions in Figure~\ref{fig:example_climb}, where we included an ablation on distributions with our R2D2-based $\epsilon z$-greedy agent over six games, finding remarkable consistency with the results of the chain MDP. We include full details and per-game figures for these ablations in the Appendix.


\begin{table}[]
\centering
\begin{tabular}{l|lll}
Algorithm (@$30$B)      & Median                       & Mean           & Human-gap   \\
\hline
\hline
R2D2                        & 16.44          & 39.55          & 0.102  \\
R2D2+RND                    & 11.14          & \textbf{42.17} & 0.151   \\
R2D2+Bootstrap    & 16.62 & 37.69          & 0.096     \\ 
R2D2+$\epsilon z$-greedy    & \textbf{17.77} & 40.16          & \textbf{0.077}     \\ 
\hline
Algorithm (@$200M$)   &                        &            &                     \\
\hline
\hline
Rainbow                     &   2.03                        &  8.82          &  0.139             \\
Rainbow+$\epsilon$-Greedy   &   2.26                       &   9.16          &  0.143       \\
Rainbow+CTS                 &   1.73                      &    6.67         & 0.150                \\
Rainbow+$\epsilon z$-Greedy  &   \textbf{2.30}              &   \textbf{9.34}  &  \textbf{0.130}         
\end{tabular}
\caption{Atari-57 final performance summaries. R2D2 results are after $30$B environment frames, Rainbow after $200$M frames.}
\label{tab:atari57}
\vspace{-10px}
\end{table}
\textbf{}
\section{Discussion and Conclusions}
We have proposed temporally-extended $\epsilon$-greedy, a form of random exploration performed by sampling an option and following it until termination, with a simple instantiation which we call $\epsilon z$-greedy. We showed, across domains and algorithms spanning tabular, linear and deep reinforcement learning that $\epsilon z$-greedy improves exploration and performance in sparse-reward environments with minimal loss in performance on easier, dense-reward environments. Further, we showed that compared with other exploration methods (pseudo-counts, RND, Bootstrap), $\epsilon z$-greedy has comparable performance averaged over the hard-exploration games in Atari, but without the significant loss in performance on the remaining games. Although action-repeats have been a part of deep RL algorithms since DQN \cite{mnih2015human}, and have been considered as a type of option \cite{schoknecht2002speeding,schoknecht2003reinforcement,vafadost2013temporal,braylan2015frame,lakshminarayanan2017dynamic,sharma2017learning}, their use for exploration with sampled durations does not appear to have been studied before.

\paragraph{Generality and Limitations}
Both $\epsilon$- and $\epsilon z$-greedy are guaranteed to converge, eventually, in the finite-state-action case, but they place probability mass over exploratory trajectories very differently, thus encoding different inductive biases. We expect there to be environments where $\epsilon z$-greedy significantly under-performs $\epsilon$-greedy. Indeed, these are easy to imagine: DeepSea with action effects randomized per-state (see Appendix Figure~\ref{fig:advdeepsea}), GridWorld with many obstacles that immediately end the episode (`mines'), a maze changing direction every few steps, etc. More generally, the limitations of $\epsilon z$-greedy are:
\textbf{(i)} Actions may not homogeneously (over states) correspond to a natural notion of shortest-path directions in the MDP.
\textbf{(ii)} Action spaces may be biased (e.g. many actions have the same effect), so that uniform action sampling may produce undesirable biased drift through the MDP.
\textbf{(iii)} Obstacles and dynamics in the MDP can cause long exploratory trajectories to waste time (e.g.\ running into a wall for thousands of steps), or produce other uninformative transitions (e.g.\ end of episode, death).

These limitations are precisely where we believe future work is best motivated. How can an agent learn stationary, problem-specific notions of direction, and learn to explore in that space efficiently? How to avoid wasteful long trajectories, perhaps by truncating early? We mentioned that this form of exploration bears similarity to the L\'evy-flights model of foraging, where an animal will abruptly end their foraging as soon as food is within sight \cite{reynolds2018current}. Could we use discrepancies in value along a trajectory to similarly truncate exploration early? Some recent work around learning action representations appear to be promising directions for investigation \cite{tennenholtz2019natural,chandak2019learning}.

Even more broadly, we believe that the successes of $\epsilon z$-greedy motivate further work on exploration directly in the space of options and hierarchical behaviours \cite{kulkarni2016hierarchical,eysenbach2018diversity}, independent of value. 

\clearpage

\bibliography{references}
\bibliographystyle{icml2020}

\clearpage
\appendix
\onecolumn

\section*{APPENDICES}

\section{Domain specifications}

\begin{figure}[t]
    \centering
    \includegraphics[width=\textwidth]{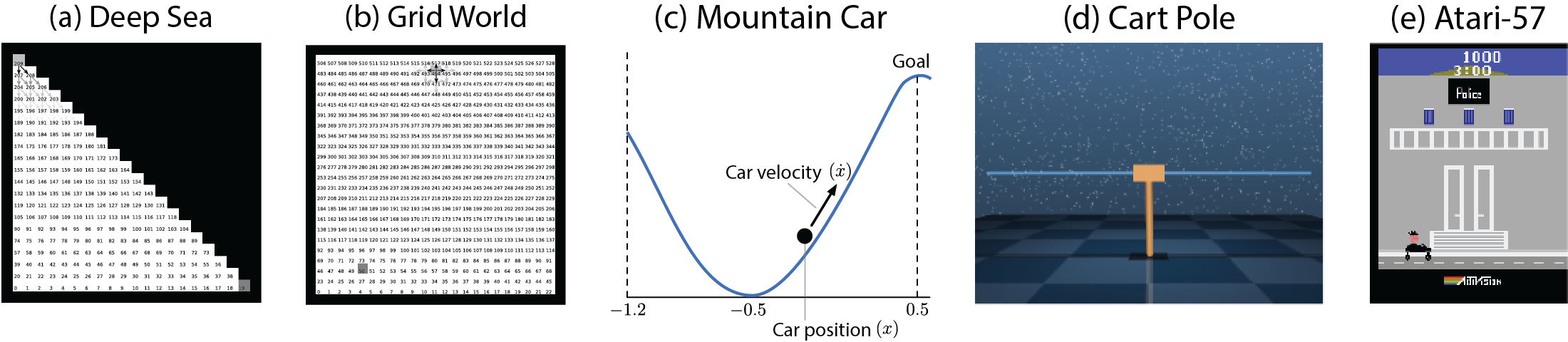}
    \vspace{-10px}
    \caption{Environments used in this work: \textbf{(a)} DeepSea, \textbf{(b)} GridWorld, \textbf{(c)} MountainCar, \textbf{(d)} CartPole, \textbf{(e)} Atari-57.}
    \label{fig:domains}
\end{figure}

\paragraph{DeepSea \cite{osband2018randomized}}
Parameterized by problem size $N$, this environment can be viewed as the lower triangle of an $N \times N$ gridworld with two actions: ``down'' and ``down-right'' which move either straight down or diagonally down and right. There is a single goal state in the far bottom-right corner, which can only be reached through a single action-sequence. The goal reward is $1.0$, and there is a per-step reward of $-0.01 / N$. Finally, all episodes end after exactly $N$ steps, once the agent reaches the bottom. Therefore, the maximum possible undiscounted return is $0.99$. An example with $N = 20$ is shown in Figure~\ref{fig:domains}a. Average first-passage times are shown for a problem size of $N = 20$ in Figure~\ref{fig:toy}a, and unlike other plots are logarithmically scaled, $\log(\mathbb{E} \left[\text{fpt} \right] + 1)$ with contour levels in the range $[0, 16]$.

In this work we use the deterministic variant of DeepSea; however, a stochastic version also exists which \textit{randomizes} the action effects at every state. That is, ``down'' may correspond to action index $0$ in one state and $1$ in another, and these assignments are performed randomly for each training run (consistently across episodes). We briefly mention this variant in our conclusions as an example in which our proposed method should be expected to perform poorly. Indeed, in Figure~\ref{fig:advdeepsea} we show that such an adversarial modification reduces $\epsilon z$-greedy's performance back to that of $\epsilon$-greedy.

For experiments, we used Q-learning with a tabular function approximator, learning rate $\alpha = 1.0$, and $\epsilon = 1.0/(N+1)$ for problem size $N$. Experiment results are averages over $5$ random seeds.

\paragraph{GridWorld}
Shown in Figure~\ref{fig:domains}b, this is an open single-room gridworld with four actions (``up'', ``down'', ``left'', and ``right''), and a single non-zero reward at the goal state. The initial state is in the top center of the grid (offset from the wall by one row), and the goal state is diagonally across from it at the other end of the room. Notice that if the goal were in the same row or column, as well as if it were placed directly next to a wall, this could be argued to advantage an action-repeat exploration method. Instead, the goal location was chosen to be harder for $\epsilon z$-greedy to find (offset from wall, far from and not in same row/column as start state).

For experiments, we used Q-learning with a tabular function approximator, learning rate $\alpha = 0.1$, $\epsilon = 0.1$, and maximum episode length $1000$. Experiment results are averages over $30$ random seeds.

Figure~\ref{fig:intro} shows average first-passage times on a similar gridworld, but with a fixed greedy policy which takes the ``down'' action deterministically.

\paragraph{MountainCar \cite{sutton2018reinforcement}}
This environment models an under-powered car stuck in the valley between two hills. The agent must build momentum in order to reach the top of one hill and obtain the goal reward. In this version of the domain all rewards are zero except for the goal, which yields reward of $1.0$. There are two continuous state variables, corresponding to the agent location, $x$, and velocity, $\dot{x}$. 

The dense-reward version of this environment can be solved reliably in less than a dozen episodes using linear function approximation on top of a low-order Fourier basis \cite{konidaris2011value}. 

In our experiments using the sparse-reward variant of the environment, we used SARSA($\lambda$) with a linear function approximation on top of an order $5$ Fourier basis. We used learning rate $\alpha = 0.005$, $\epsilon = 0.05$, $\gamma = 0.99$, and $\lambda = 0.9$. The maximum episode length was set to $5000$. Experiment results are averages over $30$ random seeds.

\paragraph{CartPole \cite{barto1983neuronlike}}
We use the ``swingup\textunderscore sparse'' variant as implemented in \citet{tassa2018deepmind}. In this sparse reward version of the environment, the agent receives zero reward unless $|x| < 0.25$ and $\cos(\theta) > 0.995$, for the cart location $x$ and pole angle $\theta$. All episodes run for $1000$ steps, and observations are $5$-dimensional continuous observation.

For experiments, we used SARSA($\lambda$) with a linear function approximation on top of an order $7$ Fourier basis. We used learning rate $\alpha = 0.0005$, $\epsilon = 0.01$, $\gamma = 0.99$, and $\lambda = 0.7$. The maximum episode length was $1000$. Weights were initialized randomly from a mean-zero normal distribution with variance $0.001$. Experiment results are averages over $30$ random seeds.

\paragraph{Atari-57}
\cite{bellemare2013arcade}, is a benchmark suite of $57$ Atari 2600 games in the Arcade Learning Environment (ALE). Observations are $210\times 160$ color images (following \cite{mnih2015human}, in many agents these are down-scaled to $84 \times 84$ and converted to grayscale). Throughout this work we use the original ALE version of Atari 2600 games, which does not include subsequently added games (beyond the 57) or features such as ``sticky actions''.

Many existing results on Atari-57 report performance of the best agent throughout training, or simply the maximum evaluation performance attained during training. We do not report this metric in the main text because it does not reflect the true learning progress of agents and tends to reflect an over estimate. However, for comparison purposes, ``best'' performance is included later in the Appendix. In the next section, alongside other agent details, we will give hyper-parameters used in the Atari-57 experiments. An example frame from the game \textsc{Private Eye} is shown in Figure~\ref{fig:domains}e.

\section{Agent and Algorithm Details}

Except for ablation experiments on the duration distribution, all $\epsilon z$-greedy experiments use a duration distribution $z(n) \propto n^{-\mu}$ with $\mu = 2.0$. These durations were capped at $n \le 10000$ for all experiments except for the Rainbow-based agents which were limited to $n \le 100$, but in this case no other values were attempted.

\subsection{Pseudo-code}

\begin{algorithm}
    \begin{algorithmic}[1]
        \Function{ezGreedy}{$Q$, $\epsilon$, $z$}
            \State $n \leftarrow 0$
            \State $\omega \leftarrow -1$
            \While{True}
                \State Observe state $x$
                \If{$n == 0$}
                    \If{$\text{random}() \le \epsilon$}
                        \State Sample duration: $n \sim z$
                        \State Sample action: $\omega \sim U(\actionspace)$
                        \State Assign action: $a \leftarrow \omega$
                    \Else
                        \State Greedy action: $a \leftarrow \arg\max_a Q(x, a)$
                    \EndIf
                \Else
                    \State Assign action: $a \leftarrow \omega$
                    \State $n \leftarrow n - 1$
                \EndIf
                \State Take action $a$
            \EndWhile
        \EndFunction
    \end{algorithmic}
    \caption{$\epsilon z$-Greedy exploration policy}
    \label{alg:ezgreedy}
\end{algorithm}

\subsection{Network Architecture.}

\textbf{Rainbow-based agents} use an identical network architecture as the original Rainbow agent \cite{hessel2018rainbow}. In particular, these include the use of NoisyNets \cite{fortunato2017noisy}, with the exception of Rainbow-CTS, which uses a simple dueling value network like the ``no noisy-nets'' ablation in \cite{hessel2018rainbow}. A preliminary experiment showed this setting with Rainbow-CTS performed slightly better than when NoisyNets were included.

\textbf{R2D2-based agents} use a slightly enlarged variant of the network used in the original R2D2 \citep{kapturowski2018recurrent}, namely a 4-layer convolutional neural network with layers of 32, 64, 128 and 128 feature planes, with kernel sizes of 7, 5, 5 and 3, and strides of 4, 2, 2 and 1, respectively. These are followed by a fully connected layer with $512$ units, an LSTM with another $512$ hidden units, which finally feeds a dueling architecture of size $512$ \citep{wang2015dueling}. Unlike the original R2D2, Atari frames are passed to this network without frame-stacking, and at their original resolution of $210 \times 160$ and in full RGB. Like the original R2D2, the LSTM receives the reward and one-hot action vector from the previous time step as inputs.

\subsection{Hyper-parameters and Implementation Notes}

Unless stated otherwise, hyper-parameters for our Rainbow-based agents follow the original implementation in \cite{hessel2018rainbow}, see Table \ref{tab:hyper_rainbow}. An exception is the Rainbow-CTS agent, which uses a regular dueling value network instead of the NoisyNets variant, and also makes use of an $\epsilon$-greedy policy (whereas the baseline Rainbow relies on its NoisyNets value head for exploration). The $\epsilon$ parameter follows a linear decay schedule $1.0$ to $0.01$ over the course of the first $4$M frames, remaining constant after that. Evaluation happens with an even lower value of $\epsilon=0.001$. The same $\epsilon$-schedule is used in Rainbow+$\epsilon$-greedy and Rainbow+$\epsilon z$-greedy, \textit{on top} of Rainbow's regular NoisyNets-based policy.

The CTS-based intrinsic reward implementation follows \cite{bellemare2016unifying}, with the scale of intrinsic rewards set to a lower value of $0.0005$. This agent was informally tuned for better performance on hard-exploration games: Instead of the ``mixed Monte-Carlo return'' update rule from \cite{bellemare2016unifying}, Rainbow-CTS uses an $n$-step Q-learning rule with $n=5$ (while the baseline Rainbow uses $n=3$), and differently from the baseline does not use a target network.

All of our R2D2-based agents are based on a slightly tuned variant of the published R2D2 agent \citep{kapturowski2018recurrent} with hyper-parameters unchanged, unless stated otherwise - see Table \ref{tab:hyper}. Instead of an $n$-step Q-learning update rule, our R2D2 uses expected SARSA($\lambda$) with $\lambda = 0.7$ \citep{van2009theoretical}. It also uses a somewhat shorter target network update period of $400$ update steps and the higher learning rate of $2\times 10^{-4}$. For faster experimental turnaround, we also use a slightly larger number of actors ($320$ instead of $256$).

The RND agent is a modification of our baseline R2D2 with the addition of the intrinsic reward generated by the error signal of the RND network from \cite{burda2018exploration}. The additional networks (``predictor'' and ``target'' in the terminology of \cite{burda2018exploration}) are small convolutional neural networks of the same sizing as the one used in \cite{mnih2015human}, followed by a single linear layer with output size $128$. The predictor is trained on the same replay batches as the main agent network, using the Adam optimizer with learning rate $0.0005$. The intrinsic reward derived from its loss is normalized by dividing by its variance, utilizing running estimates of its empirical mean and variance. Note, the RND agent includes the use of $\epsilon$-greedy exploration.

The Bootstrapped R2D2 agent closely follows the details of \cite{osband2016deep}. The network is extended to have $k = 8$ action-value function heads which share a common convolutional network, but with distinct fully-connected layers on top (each with the same dimensions as in R2D2). During training, each actor samples a head uniformly at random, and follows that action-value function's $\epsilon$-greedy policy for an entire episode. Each step, a mask is sampled, and added to replay, with probability $p=0.5$ indicating which heads will be trained on that step of experience. During evaluation, we compute the average of each head's $\epsilon$-greedy policy to form an ensemble policy that is followed.

\begin{table}[h!]
    \centering
    \begin{tabular}{c|c}
         \textbf{Rainbow} (baseline) \\
         \hline
         
         Replay buffer size & $10^6$ observations \\
         Priority exponent & $0.5$ \\
         Importance sampling exponent & annealed from $0.4$ to $1.0$ over the course of $200$M frames \\ 

         Multi-step returns $n$ & 3 \\
         Discount $\gamma$ & $0.99$ \\
         Minibatch size & $32$ \\
        
         Optimiser & Adam \\
         Optimiser settings & learning rate $= 6.25 \times 10^{-5}$, $\varepsilon=1.5 \times 10^{-4}$ \\
         Target network update interval & $2000$ updates ($32$K environment frames)\\
         
         $\epsilon$ (training) & $0.0$ (i.e. no $\epsilon$-greedy used) \\ 
         $\epsilon$ (evaluation) & $0.0$ (i.e. no $\epsilon$-greedy used) \\ 
         
         \hline \\
         \hline
         \textbf{Rainbow+}$\epsilon$/$\epsilon z$\textbf{-greedy, Rainbow+CTS} \\
         \hline
         $\epsilon$ (training) & \textbf{linear decay from 1.0 to 0.01 over the course of 4M frames} \\
         $\epsilon$ (evaluation) & \textbf{0.001} \\ 
         
         \hline \\
         \hline
         \textbf{Rainbow+CTS only} \\
         \hline
         Multi-step returns $n$ & \textbf{5} \\
         Intrinsic reward scale ($\beta$ in \cite{bellemare2016unifying}) & \textbf{0.0005} \\ 
         Target network update interval & \textbf{1 (i.e., no target network used)} \\

    \end{tabular}
    \caption{Hyper-parameters values used in Rainbow-based agents (deviations from \cite{hessel2018rainbow} highlighted in boldface).}
    \label{tab:hyper_rainbow}
\end{table}

\begin{table}[h!]
    \centering
    \begin{tabular}{c|c}
         Number of actors &  $\boldsymbol{320}$  \\
         Actor parameter update interval & $400$ environment steps \\
         \hline
         
         Sequence length & $80$ (\textbf{+ prefix of 20 for burn-in}) \\ 
         Replay buffer size & $4\times 10^6$ observations ($10^5$ part-overlapping sequences) \\
         Priority exponent & $0.9$ \\
         Importance sampling exponent & $0.6$ \\ 

         \hline
         Learning rule & \textbf{Expected SARSA}($\boldsymbol{\lambda}$), $\boldsymbol{\lambda = 0.7}$ \\ 
         Discount $\gamma$ & $0.997$ \\
         Minibatch size & $64$ \\
        
         Optimiser & Adam \\
         Optimiser settings & \textbf{learning rate} $\boldsymbol{= 2\times 10^{-4}}$, $\varepsilon=10^{-3}$ \\
         Target network update interval & \textbf{400 updates} \\

    \end{tabular}
    \caption{Hyper-parameters values used in R2D2-based agents (deviations from \citep{kapturowski2018recurrent} highlighted in boldface).}
    \label{tab:hyper}
\end{table}

\section{Experiment Details}\label{app:expdetails}

\begin{table}[]
\begin{tabular}{l|lllll}
\textbf{Environment} & \textbf{\# Trials} & \textbf{\# Steps}              & \textbf{Max Episode Length} & \textbf{Contour Scale} & \textbf{Discretization} \\
\hline
DeepSea              & 5                  & 500000 $\times N$ & N                           & Log                    & None                    \\
GridWorld            & 100                & 5000                           & 5000                        & Linear                 & None                    \\
MountainCar          & 50                 & 5000                           & 5000                        & Linear                 & 12                      \\
CartPole             & 100                & 5000                           & 5000                        & Linear                 & 20                     
\end{tabular}
\caption{Settings for experiments used to generate average first-visit visualizations found in main text.}
\label{tab:firstpass}
\end{table}

\paragraph{First-visit visualizations} These results (e.g. see Figure~\ref{fig:intro}) are intended to illustrate the differences in state-visitation patterns between $\epsilon$-greedy and $\epsilon z$-greedy. These are generated with some fixed $\epsilon$, often $\epsilon = 1.0$ for pure-exploration independent of the greedy policy, and are computed using Monte-Carlo rollouts with each state receiving an integer indicating the first step at which that state was visited on a given trial. States that are never seen in a trial receive the maximum step count, and we then average these over many trials. For continuous-state problems we discretize the state-space and count any state within a small region for the purposes of visitation. We give these precise values in Table~\ref{tab:firstpass}.

\paragraph{Atari experiments}

The experimental setup for the Rainbow-based and R2D2-based agents each match those used in their respective baseline works. In particular, Rainbow-based agents perform a mini-batch gradient update every $4$ steps and every $1$M environment frames learning is frozen and the agent is evaluated for $500$K environment frames. In the R2D2-based agents, acting, learning, and evaluating all occur simultaneously and in parallel, as in the baseline R2D2.

In the Atari-57 experiments, all results for Rainbow agents are averaged over $5$ random seeds, while results for R2D2-based agents are averages over $3$ random seeds.

The human-normalized score is defined as
$$score = \frac{\text{agent} - \text{random}}{\text{human} - \text{random}},$$
where agent, random, and human are the per-game scores for the agent, a random policy, and a human player respectively. The \emph{human-gap} is defined as the average, over games, performance difference between human-level over all games,
$$human\_gap = 1.0 - \mathbb{E}\min(1.0, score).$$

\section{Further Experimental Results}

\begin{figure}[t]
    \centering
    \includegraphics[width=.5\textwidth]{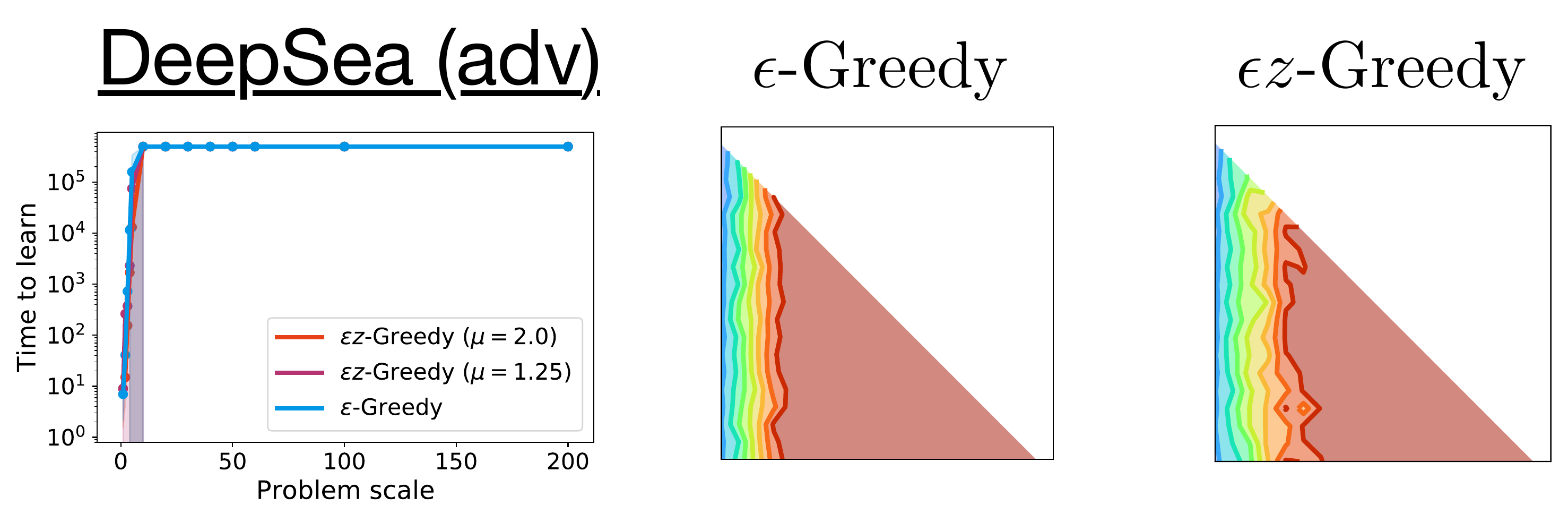}
    \vspace{-10px}
    \caption{Adversarial modification to DeepSea environment causes $\epsilon z$-greedy to perform no better than $\epsilon$-greedy.}
    \label{fig:advdeepsea}
\end{figure}

In this section, we include several additional experimental results that do not fit into the main text but may be helpful to the reader. In the conclusions we highlight a limitation of $\epsilon z$-greedy which occurs when the effects of actions differ significantly between states. In Figure~\ref{fig:advdeepsea} we present results for such an adversarial setting in the DeepSea environment, where the action effects are randomly permuted for every state. We observe, as expected, that in this setting $\epsilon z$-greedy no longer provides more efficient exploration than $\epsilon$-greedy.

\begin{figure*}[h]
    \centering
    \includegraphics[width=.55\textwidth]{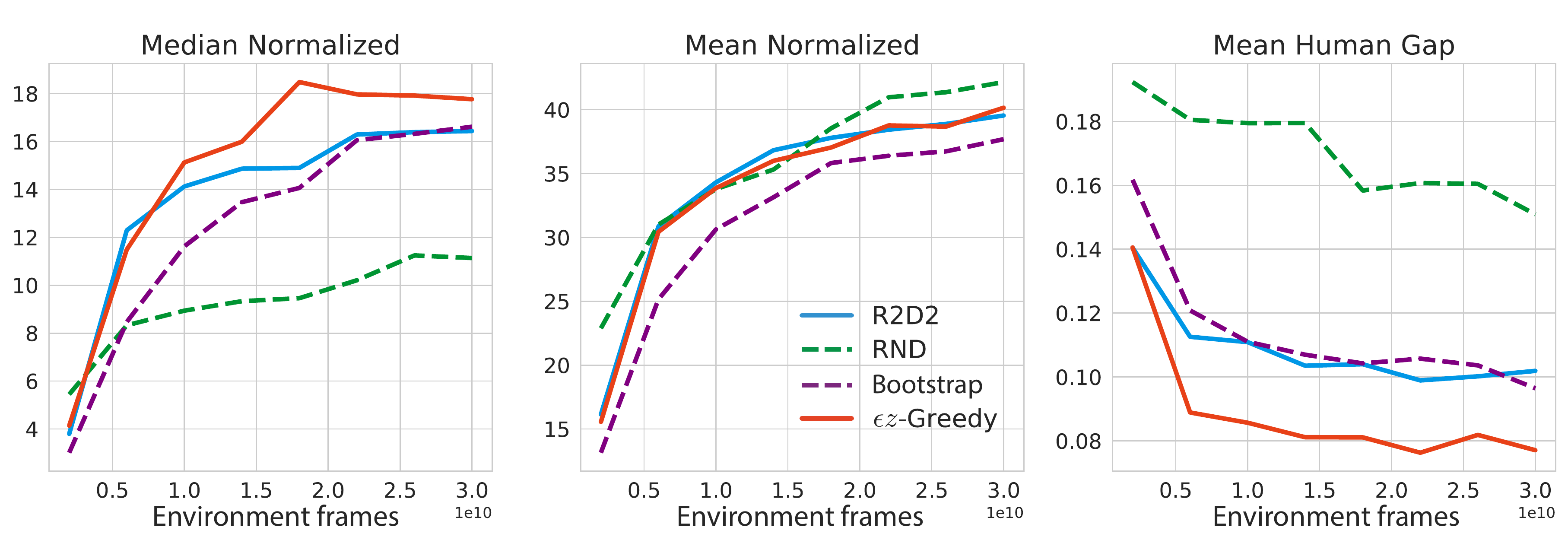}
    \includegraphics[width=.55\textwidth]{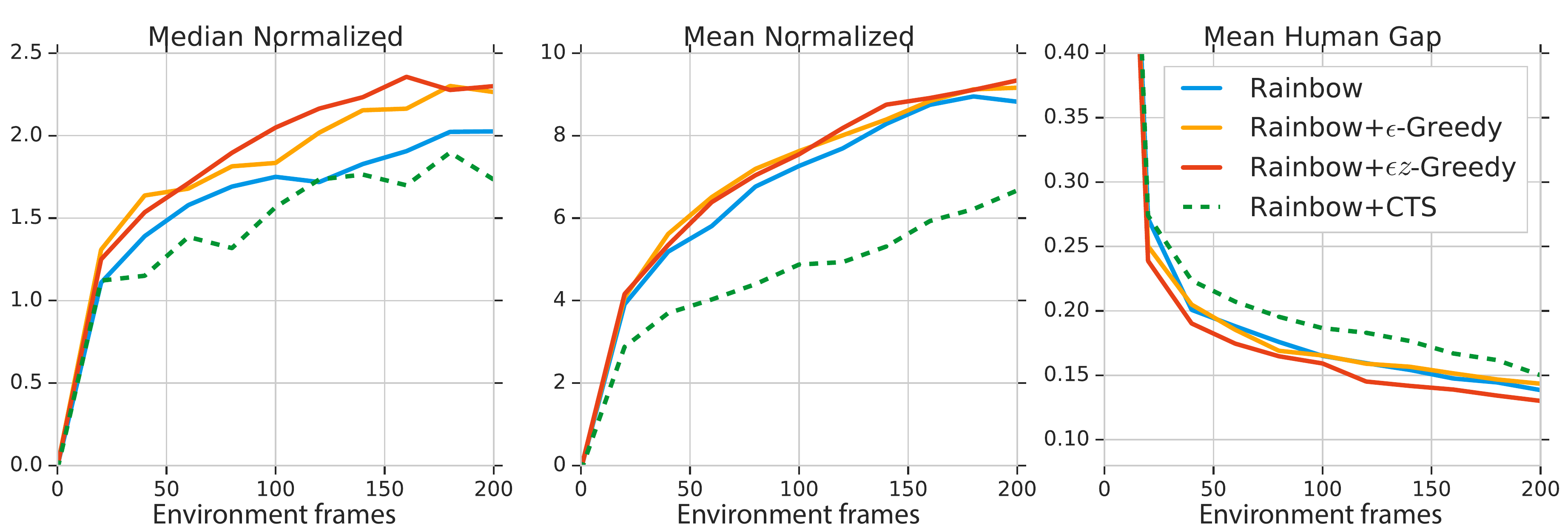}
    \caption{Atari-57 summary curves for R2D2-based methods (top) and Rainbow-based methods (bottom).}\label{fig:r2d2_atari_summary}
\end{figure*}

\begin{figure*}[h]
    \centering
    \includegraphics[width=\textwidth]{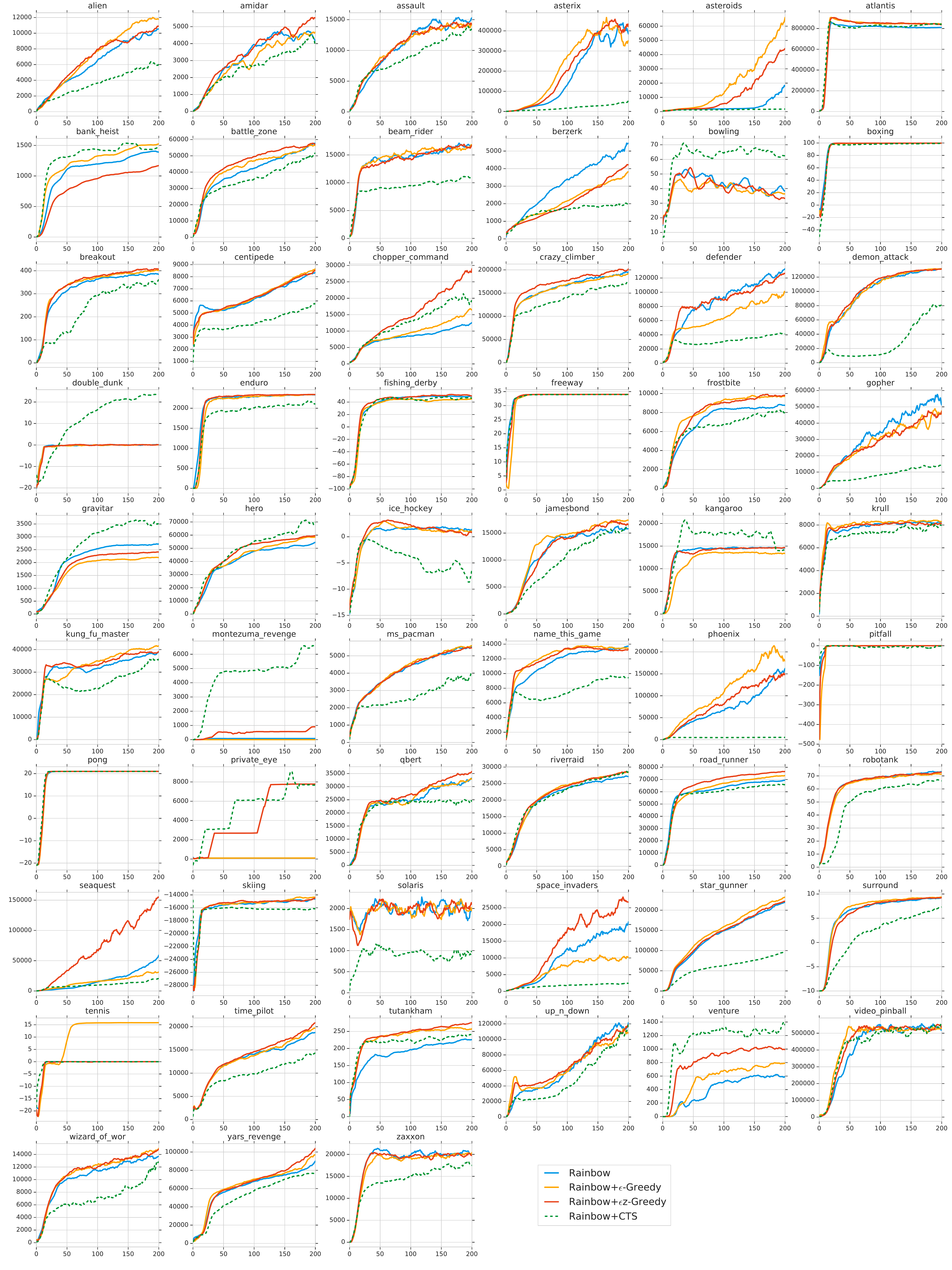}
    \caption{Per-game Atari-57 results for Rainbow-based methods.}\label{fig:rainbow_atari_full}
\end{figure*}

\begin{figure*}[h]
    \centering
    \includegraphics[width=.8\textwidth]{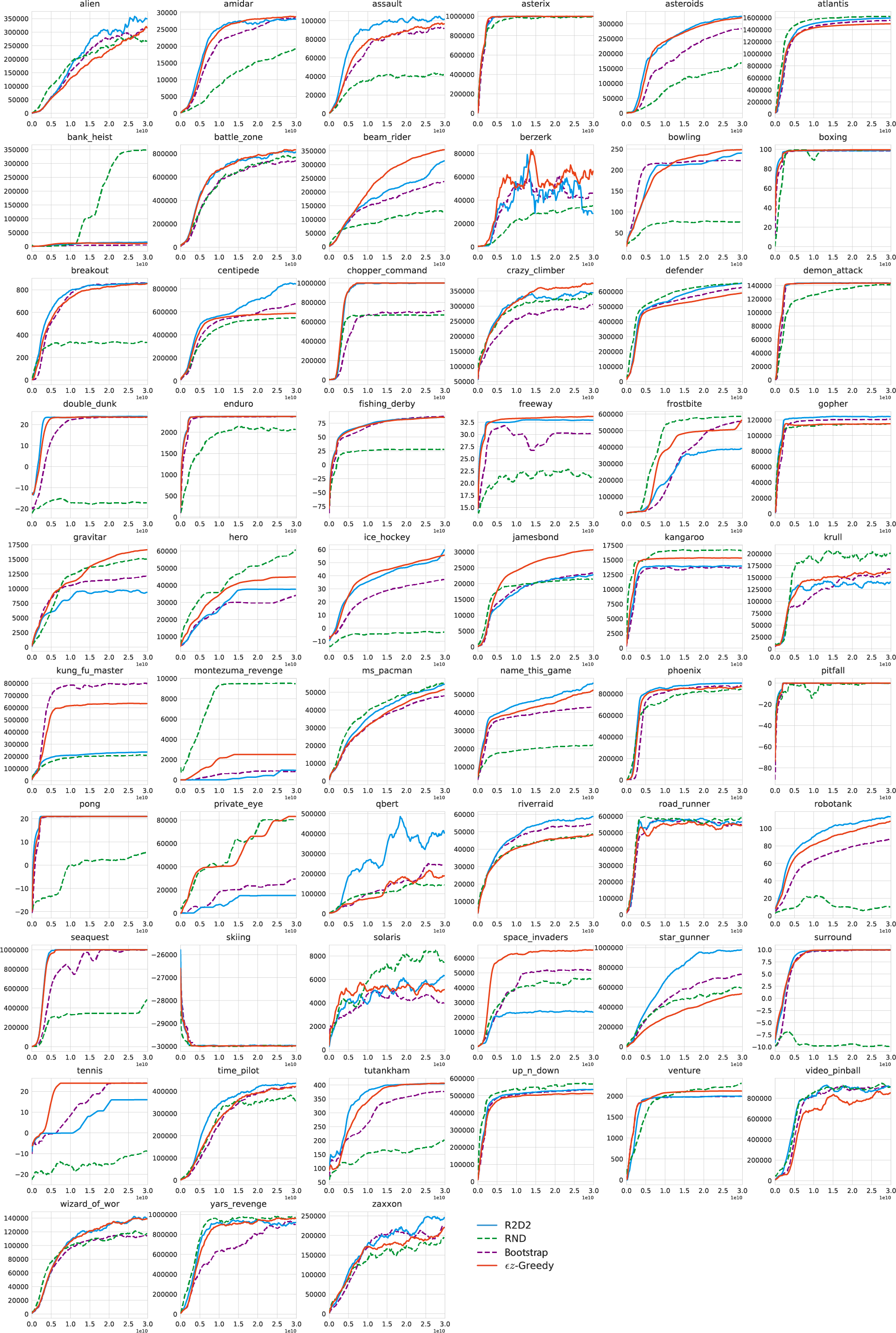}
    \caption{Per-game Atari-57 results for R2D2-based methods.}\label{fig:r2d2_atari_full}
\end{figure*}

In the main text we give summary learning curves on Atari-57 for Rainbow- and R2D2-based agents in terms of median human-normalized score and human-gap. In Figure~\ref{fig:r2d2_atari_summary} we show these as well as the mean human-normalized score learning curves. In Figures~\ref{fig:rainbow_atari_full} \&~\ref{fig:r2d2_atari_full} we give full, per-game results for Rainbow- and R2D2-based agents respectively. 

Finally, in Table~\ref{tab:atari57} we give mean and median human-normalized scores and the human-gap on Atari-57 for the final trained agents. However, this is a slightly different evaluation method than is often used \cite{mnih2015human,hessel2018rainbow}, in which only the best performance for each game, over training, is considered. For purposes of comparison we include these results in Table~\ref{tab:atari_full}.

\begin{table}[]
\centering
\begin{tabular}{l|lll|ll}
Algorithm (@$30$B)          & Median         & Mean           & Human-gap      & Median (best)   & Mean (best) \\
\hline
\hline
R2D2  & 16.44          & 39.55          & 0.102          & 19.36           & 46.98  \\
R2D2+RND & 11.14          & \textbf{42.17} & 0.151          & 14.34           & \textbf{48.02} \\
R2D2+Bootstrap  & 16.62         & 37.69 & 0.096          & 19.35           & 43.87 \\
R2D2+$\epsilon z$-greedy    & \textbf{17.77} & 40.16          & \textbf{0.077} & \textbf{22.63}  & 45.33      \\ 
\hline
Algorithm (@$200M$)   &                        &            &                     \\
\hline
\hline
Rainbow                     &   2.03                        &  8.82          &  0.139       &  2.20   &   12.24              \\
Rainbow+$\epsilon$-Greedy   &   2.26                       &   9.16          &  0.143     &  2.56  &    12.23        \\
Rainbow+CTS                 &   1.73                      &    6.67         & 0.150         &  2.09  &  7.62               \\
Rainbow+$\epsilon z$-Greedy  &   \textbf{2.30}              &   \textbf{9.34}  &  \textbf{0.130}      &  \textbf{2.74}  &   \textbf{12.28}
\end{tabular}
\caption{Atari-57 final performance summaries. R2D2 results are after $30$B environment frames, and Rainbow results are after $200$M environment frames. Note the main text only has Rainbow-based results up to $120$M frames. We also include median and mean human-normalized scores obtained by using \textit{best} (instead of \textit{final}) evaluation scores for each training run, to allow comparison with past publications which often used this metric (e.g. \cite{hessel2018rainbow}).}
\label{tab:atari_full}
\end{table}

\clearpage

\clearpage

\end{document}